\documentclass[acmtog,screen,nonacm]{acmart}

\usepackage{booktabs} 

\usepackage{amsfonts}
\usepackage{pdfpages}
\usepackage{enumitem}

\usepackage[ruled]{algorithm2e} 

\SetAlFnt{\small}
\SetAlCapFnt{\small}
\SetAlCapNameFnt{\small}
\SetAlCapHSkip{0pt}
\usepackage{subcaption}
\usepackage{booktabs}
\usepackage{multirow}
\usepackage{adjustbox}
\usepackage{xspace}
\usepackage{pifont}
\usepackage{xcolor}

\definecolor{darkred}{RGB}{204, 0, 0}
\definecolor{darkgreen}{RGB}{0, 160, 0}

\newcommand{\ME}[1]{\textcolor{blue}{[\textbf{ME:} #1]}}

\newcommand{\eg}{\textit{e}.\textit{g}.\,\@\xspace}
\newcommand{\ie}{\textit{i}.\textit{e}.\,\@\xspace}
\newcommand{\etal}{\textit{et al}.\@\xspace}
\newcommand{\etc}{\textit{etc}}

\usepackage{amsmath,amsfonts,bm}
\usepackage{mathtools}

\def\1{\bm{1}}

\DeclareMathAlphabet{\mathsfit}{\encodingdefault}{\sfdefault}{m}{sl}
\SetMathAlphabet{\mathsfit}{bold}{\encodingdefault}{\sfdefault}{bx}{n}

\newcommand{\E}{\mathbb{E}}

\newcommand{\Eb}[2]{\E_{#1}\!\left[#2\right]}

\newcommand{\bI}{\mathbf{I}}

\newcommand{\bzero}{\mathbf{0}}

\newcommand{\bc}{\mathbf{c}}
\newcommand{\bd}{\mathbf{d}}
\newcommand{\be}{\mathbf{e}}

\newcommand{\bv}{\mathbf{v}}

\newcommand{\bx}{\mathbf{x}}
\newcommand{\bzeta}{\mathbf{\zeta}}

\newcommand{\bz}{\mathbf{z}}

\newcommand{\bepsilon}{{\boldsymbol{\epsilon}}}

\newcommand{\bP}{\mathbf{P}}

\newcommand{\timenear}{t_n}
\newcommand{\timefar}{t_f}
\newcommand{\ray}{\mathbf{r}}
\newcommand{\Ctrue}{C(\ray)}
\newcommand{\expo}[1]{\exp\left(#1\right)}
\newcommand{\absrp}{\sigma}

\newcommand{\VG}[1]{{\textcolor{darkgreen}{{\textbf{[VG: #1]}}}}}

\acmJournal{TOG}

\newcommand{\OURS}{AvatarStudio\xspace} 

\begin{document}
\title{\OURS: Text-driven Editing of 3D Dynamic Human Head Avatars} 
%
\author{Mohit Mendiratta}
\affiliation{%
	\institution{Max Planck Institute for Informatics and Saarland University}
    \country{Germany}
}
\email{mmendira@mpi-inf.mpg.de}
\author{Xingang Pan}
\authornote{Indicates equal contribution}
\affiliation{%
	\institution{Max Planck Institute for Informatics, SIC}
 \country{Germany}
}
\email{xpan@mpi-inf.mpg.de}
\author{Mohamed Elgharib}
\authornotemark[1]
\affiliation{%
	\institution{Max Planck Institute for Informatics, SIC}
 \country{Germany}
}
\email{elgharib@mpi-inf.mpg.de}
\author{Kartik Teotia}
\affiliation{%
	\institution{Max Planck Institute for Informatics and Saarland University}
    \country{Germany}
}
\email{ktoetia@mpi-inf.mpg.de}
\author{Mallikarjun B R}
\affiliation{%
	\institution{Max Planck Institute for Informatics and Saarland University}
 \country{Germany}
}
\email{mbr@mpi-inf.mpg.de}
\author{Ayush Tewari}
\affiliation{%
	\institution{MIT CSAIL}
\country{United States of America}
}
\email{ayusht@mit.edu}
\author{Vladislav Golyanik}
\affiliation{%
	\institution{Max Planck Institute for Informatics, SIC}
 \country{Germany}
}
\email{golyanik@mpi-inf.mpg.de}
\author{Adam Kortylewski}
\affiliation{%
	\institution{University of Freiburg and Max Planck Institute for Informatics, SIC}
 \country{Germany}
}
\email{akortyle@mpi-inf.mpg.de}
\author{Christian Theobalt}
\affiliation{\institution{Max Planck Institute for Informatics, SIC}
\country{Germany}
 }
 \email{theobalt@mpi-inf.mpg.de}

\begin{abstract}
Capturing and editing full head performances enables the creation of virtual characters with various applications such as extended reality and media production. 
The past few years witnessed a steep rise in the photorealism of human head avatars. 
Such avatars can be controlled through different input data modalities, including RGB, audio, depth, IMUs and others. 
While these data modalities provide effective means of control, they mostly focus on editing the head movements such as the facial expressions, head pose and/or camera viewpoint. 
In this paper, we propose \OURS, a text-based method for editing the appearance of a dynamic full head avatar.
Our approach builds on existing work to capture dynamic performances of human heads using neural radiance field (NeRF) and edits this representation with a text-to-image diffusion model.
Specifically, we introduce an optimization strategy for incorporating multiple keyframes representing different camera viewpoints and time stamps of a video performance into a single diffusion model.
Using this personalized diffusion model, we edit the dynamic NeRF by introducing view-and-time-aware Score Distillation Sampling (VT-SDS) following a model-based guidance approach.
Our method edits the full head in a canonical space, and then propagates these edits to remaining time steps via a pretrained deformation network.
We evaluate our method visually and numerically via a user study, and results show that our method outperforms existing approaches.Our experiments validate the design choices of our method and highlight that our edits are genuine, personalized, as well as 3D- and time-consistent.
\end{abstract}

%
%


%
%

\begin{teaserfigure}
 \centering
	\includegraphics[width=1.0\textwidth]{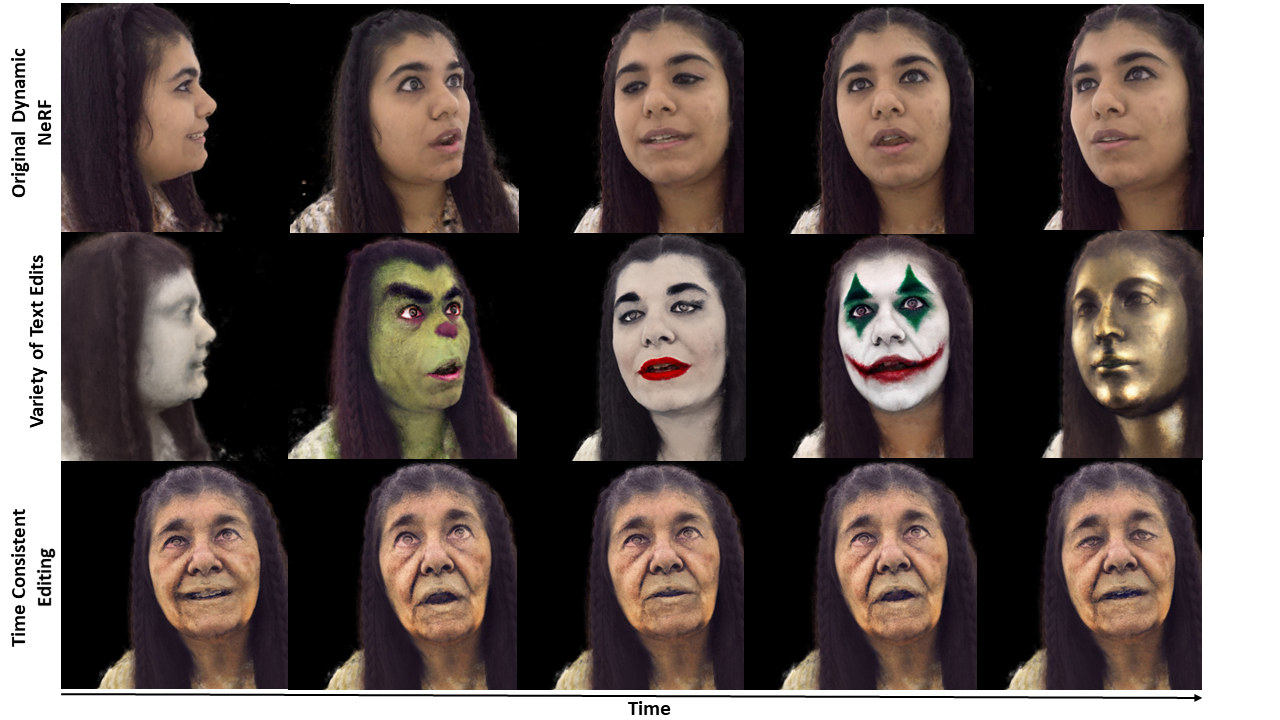}
	\caption{Our method AvatarStudio takes as input a 3D NeRF volume of a dynamic head (top) and produces visual edits that correspond to a target text prompt (second and third rows). Our method is the first designed specifically to handle text-based editing of videos. It also produces 3D-consistent results that can be viewed from an arbitrary camera viewpoint.
    }	\label{fig:teaser}
\end{teaserfigure}

\maketitle

\section{Introduction} \label{sec:intro}
The human face is at the center of our visual communications and hence its digitization is of utmost importance. 
The past few years have witnessed a sharp rise in the photorealism of digital faces. 
To achieve this, several methods were proposed, such as generative adversarial networks for 2D images~\cite{StyleGAN,ProgressiveGANs}. 
Other methods build such high-quality in 3D using either explicit~\cite{DAM,gecer2021fast}, or more recently, learnable implicit~\cite{MVP,IMAvatar} scene representations.
In addition to photorealism, controlling and rigging digital faces received a lot of attention. 
This includes for instance methods \cite{grassal2021neural,athar2022rignerf,Gao2022nerfblendshape}
that utilize a low-dimensional parametric representation in form of 3D morphable model~\cite{blanz1999morphable,egger20203d} or some other latent space~\cite{teotia2023hq3davatar,StyleFlow,Wang_2021_CVPR}. 
Moreover, several data modalities have been explored as a control signal, such as RGB images~\cite{Siarohin19,Recycle-GAN,MVP}, audio~\cite{SynObama,thies2020nvp}, sparse image representations such as contours and keypoints~\cite{zakharov2019fewshot,KeyPointNeRF} and even input from sensors such as IMUs and IR cameras~\cite{li2015facial,Wei19}.  

The vast majority of existing methods for controlling digital faces, however, focus on editing the motion of the face. 
That is, controlling the facial expressions, head pose and/or the camera viewpoint~\cite{kirschstein2023nersemble,raj2021pva}. 
Controlling the facial appearance has been mostly studied in the context of facial relighting~\cite{rao2022vorf,neural3drelightable,tan2022voluxgan}. 
Here, methods were developed that edit the facial appearance as a function of the scene illumination. \
For this, the target illumination is commonly described via HDRI maps~\cite{SIPR,mallikarjun2021photoapp} or via a low dimensional representation such as spherical harmonics~\cite{DIPR,tewari2020pie,neural3drelightable}. 
There are also methods for editing the facial appearance in a non-photorealistic manner~~\cite{selim16,Fiser17SIG,yang2022Vtoonify}. These methods usually take a target painting as input and can handle moving heads.  
One input modality that has not been fully explored yet for facial edits is text.
Text is one of the most user-friendly data modalities that can be easily defined without any expert knowledge. 

In the past few years, text-driven image synthesis attracted the attention of the research community. Thanks to the wide development and adaptation of transformers~\cite{AttentionAllYouNeed,Radford2018ImprovingLU} and diffusion models~\cite{rombach2021highresolution}, several works have shown the ability of editing images in 2D~\cite{brooks2022instructpix2pix,ruiz2022dreambooth} and 3D~\cite{poole2022dreamfusion,instructnerf2023,jain2021dreamfields}, given text prompt as input. While 2D-based methods produce interesting results, they can not produce edits that are 3D-consistent. In contrast, 3D-based methods~\cite{instructnerf2023,poole2022dreamfusion,jain2021dreamfields} show results on a 3D volume that can be rendered faithfully from an arbitrary camera viewpoint. However, even such methods lack in several ways. First, many of them~\cite{jain2021dreamfields,aneja2022clipface,wang2021clip,wang2022nerf} optimize their solution in a joint image-text embedding known as CLIP~\cite{Radford2018ImprovingLU}. While such CLIP-based objective function leads to interesting results, it usually tends to generate limited edits. Second, none of the existing methods are designed to handle dynamic scenes and hence cannot process image sequences properly. This usually leads to clear artifacts and limited edibility (as shown in our experiments). 

In this paper, we propose \OURS, a text-based method for editing the appearance of a dynamic full head avatar. 

Our approach assumes a digital head avatar as input, that can be trained from a multi-view performance capture of a human head.
In particular, we follow the approach presented in HQ3DAvatar\cite{teotia2023hq3davatar} to learn a volumetric head avatar as being one of the latest in literature. 
Here, the head is represented as a canonical neural radiance field (NeRF) \cite{mildenhall2021nerf} and a deformation network propagates the canonical representation across time.
Our approach enables the text-based editing of such dynamic volumetric avatars in a view- and time-coherent manner.

Specifically, we perform editing through text-based conditional image generation with a diffusion model.
We make several technical contributions to ensure that the editing of \OURS are genuine, personalized, as well as 3D- and time-consistent.
First, we sample several keyframes from the multi-view video that represent different camera viewpoints and time stamps of the performance capture.
We introduce an optimization strategy to incorporate these keyframes into a single diffusion model, by fine-tuning a pre-trained model with a unique text identifier as proposed in \cite{ruiz2022dreambooth}.
Importantly, to prevent the leakage of information between keyframes we keep the sampled noise constant for every batch during each fine-tuning iteration.
Based on this personalized diffusion model we can generate and edit each keyframe individually. 
We leverage this property to edit the dynamic NeRF
by introducing a novel view- and time-aware Score Distillation Sampling (VT-SDS) approach, that iteratively edits the dynamic NeRF by sampling a random set of keyframes across the view and time domain. 
VT-SDS follows a model-based classifier-free guidance approach \cite{zhang2022sine}, where we take advantage of the step-by-step generation process in diffusion models to guide the early stages of the image generation towards the content of the respective keyframe, while performing the editing throughout the later stages of the generation process with a large-scale pre-tained diffusion model. 
To ensure that the edited dynamic neural radiance field remains
faithful and free from overfitting artifacts we use an annealing strategy that gradually lowers the effect of the personalized diffusion model to enable high-frequency edits.
Aspects of novelty of this work include:
\begin{itemize}
    \item We present the first method for text-driven editing of dynamic 3D human head avatars. Our approaches leverages the state-of-the-art in neural volumetric scene representations together with recent advances in text-driven diffusion models 
    to achieve high-quality editing of dynamic digital avatars.   
    \item A new optimization strategy for incorporating multiple keyframes that represent different camera viewpoints and different time stamps, into a single diffusion model. 
    \item A view- and time-aware Score Distillation Sampling (VT-SDS) that enables high-quality personalized editing in a coherent manner across the view and time domain.
\end{itemize}
We evaluate our method subjectively and numerically through a user study and compare against related methods. Results show that our approach produces a wide variety of text-based edits, while maintaining 
the integrity of the input identity (see Fig.~\ref{fig:teaser}). It generates temporally coherent results and clearly outperforms related methods.

\section{Related Work}
This section provides an overview of generative models for image synthesis, with emphasize on diffusion models. First, we introduce Generative Adversarial Networks (GANs) and outline some of the landmark works in the literature~\cite{ProgressiveGANs,StyleGAN}. We then discuss diffusion models in details, and focus on methods that can edit images using text as input. Here, diffusion models are divided into two main categories; 2D~\cite{ruiz2022dreambooth,brooks2022instructpix2pix} and 3D~\cite{wang2021clip,wang2022nerf,jain2021dreamfields,poole2022dreamfusion,instructnerf2023}. One main difference between both approaches is that 3D methods can produce edits that are multi-view consistent, while 2D methods do not focus on changing the camera viewpoint. We outline important milestones in diffusion models for image synthesis. This includes enabling object-specific edits as in DreamBooth~\cite{ruiz2022dreambooth}, and the introduction of the probability density distillation in DreamFusion~\cite{poole2022dreamfusion}. We also discuss the CLIP image-text embedding~\cite{CLIP}, and how it is commonly used in the literature in formulating the objective function~\cite{jain2021dreamfields,aneja2022clipface,wang2021clip,wang2022nerf}. Our work differs from related works in several ways. To start with, it is the first that enables text-driven editing of image sequences. This is done by the introduction of a novel optimization that allows incorporating temporal frames in a pre-trained diffusion model. Second, our method is 3D by design, thus enables edits that are multi-view consistent. Last by not least, we do not use the CLIP embedding and thus enable stronger and more faithful edits. We now discuss related methods in more details.
\subsection{Generative Models for Image Synthesis}
The use of Generative Adverserial Networks, or GANs, for image synthesis have been an active research topic for the past years. This goes back to the revolutionary work of Goodfellow~\etal~\cite{GANGoodfellow}, 
where it was shown that a generator and a discriminator can be trained in an adversarial manner until the discriminator is no longer capable of telling whether the generator's output is real or synthesized. 
Results showed the ability of GANs to synthesis low resolution images of faces and other objects. This sparked a plethora of follow up work in generative models for image synthesis, including
two notable works, Progressive GANs~\cite{ProgressiveGANs} and StyleGAN~\cite{StyleGAN}. Here, GANs were able to synthesize high resolution images (1K) with extreme photorealism. 

It was not late until diffusion models found their way into image synthesis. To this end, Ho~\etal~\cite{ho2020denoising} have shown that diffusion probabilistic models (DPM)
can be represented as a Markov Chain process using autoencoders. Briefly after in Dhariwal~\etal~\cite{Dhariwal21}, it was shown that diffusion models can beat GANs in image synthesis quality. 
However, one main concern still remained; the computational complexity of such models. Rombach~\etal~\cite{rombach2021highresolution} addressed this concern 
by training diffusion models on latent spaces of autoencoders. Furthermore, several means of conditioning the diffusion model were shown, including text. This sparked a greater interest in text-driven image synthesis, especially with the rising popularity of transformers~\cite{AttentionAllYouNeed,Radford2018ImprovingLU}. While earlier works of text-driven synthes such as DALL-E~\cite{ramesh2021zeroshot} relied primarily on transformers for language modeling and image synthesis, the follow-up version DALL-E 2~\cite{ramesh2022hierarchical} utilized diffusion models. Other text-driven synthesis methods were also proposed using other generative models such as StyleGAN~\cite{ramesh2021zeroshot}.
However, with the availability of public implementations of diffusion models such as Stable Diffusion~\cite{StableDiff,rombach2021highresolution}, text-driven image synthesis have witnessed great progress in the past couple of years. 

\subsection{Text-driven Diffusion Models for Image Synthesis}
Current methods for text-driven image synthesis using diffusion models can be classified into 2D~\cite{ruiz2022dreambooth,brooks2022instructpix2pix} and 3D~\cite{wang2021clip,wang2022nerf,jain2021dreamfields,aneja2022clipface,poole2022dreamfusion,lin2022magic3d,instructnerf2023} approaches. While the former produces a wide range of visual edits in terms of content and style, the latter produces results that are 3D-consistent and thus can be viewed from an arbitrary camera angle. DreamBooth~\cite{ruiz2022dreambooth} is a 2D-based approach that handles the problem of fine-tuning large text-to-image diffusion models to a specific examined object. Here, multiple images (typically 3-5) of the same object is provided as input, while  DreamBooth learns to associate a unique identifier to this object. This embeds the examined object in the output domain of the text-to-image diffusion model, thus allowing a wide variety of text-driven edits.
Instruct-Pix2Pix~\cite{brooks2022instructpix2pix} takes a different approach for the same problem of text-driven image synthesis. Their idea is to generate paired synthetic data by utilizing the large language model of GPT-3~\cite{GPT3}
together with the text-to-image model of Stable Diffusion. This strategy generalizes well to real user-instructions and real input images during test.
Unlike our method, none of these 2D-based methods can generate results that are multi-view consistent~\cite{ruiz2022dreambooth,brooks2022instructpix2pix}.  

3D text-driven image synthesis methods can be classified as ones that utilize a CLIP embedding~\cite{jain2021dreamfields,aneja2022clipface,wang2021clip,wang2022nerf} and others that use other means for optimizing their solution~\cite{poole2022dreamfusion,lin2022magic3d,instructnerf2023}. CLIP~\cite{CLIP}, short for "Contrastive Language-Image Pre-Training", is a joint text and image embedding trained in a way to predict the correct (text,image) pairing. 
Using CLIP and a Neural Radiance Field (NeRF)~\cite{mildenhall2021nerf} formulation, Dream Fields~\cite{jain2021dreamfields} extended 2D text-to-diffusion models to 3D. 
Along similar lines, DreamFusion~\cite{poole2022dreamfusion} also extended 2D models to 3D, however not using CLIP embedding. Instead, a new loss is proposed based on probability density distillation.
Here, a solution is initialized with a random 3D NeRF model, which is then optimized in a way so that 2D renderings from an arbitrary viewpoint minimizes the loss.  
Magic3D~\cite{lin2022magic3d} improves upon the computational efficiency of DreamFusion using a coarse-to-fine manner. Following the 2D text-to-image model of Instruct-Pix2Pix~\cite{brooks2022instructpix2pix}, Instruct-NeRF2NeRF~\cite{instructnerf2023}
proposes a 3D text-to-image solution. Here, 2D images are iteratively edited using  Instruct-Pix2Pix and the resulting 3D NeRF model is iteratively optimized. Results show 3D-consistent edits of various forms without the need of any 
additional training data. 

ClipFace~\cite{aneja2022clipface} is a self-supervised approach for text-driven editing of human faces. Here, the face is modelled through a 3D morphable model (3DMM), where each facial component is trained separately. The solution uses a combination of a CLIP-based loss together with an adversarial training. 
While CLIP based methods achieve good results, they still can be limited in terms of their edibility. In addition, using a 3DMM lacks the ability of editing the full head~\cite{aneja2022clipface}, and thus discard important regions such as the mouth interior. Finally, none of the remaining 3D-based methods are designed to handle image sequences and thus generate clear artifacts with limited editing, in contrast to our approach. Last but not least, we would like to point out that many of these methods are concurrent work~\cite{aneja2022clipface,instructnerf2023}. Despite that, we still compare against the most related methods to us~\cite{instructnerf2023,jain2021dreamfields,poole2022dreamfusion}.

\begin{figure*}
    \centering    \includegraphics[width=\textwidth]{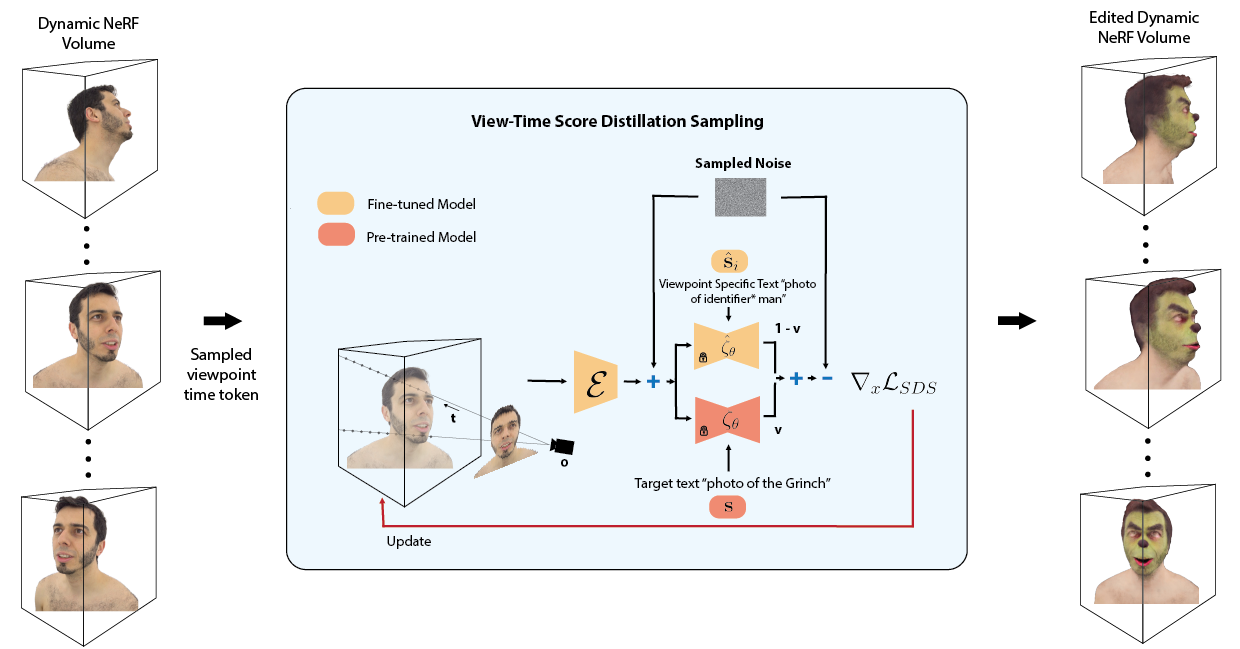}
    \caption{\textbf{An overview of our method of our approach for text-driven editing of dynamic head avatars.} Our method takes as input a reconstructed dynamic NeRF volume (left) and a text prompt \textbf{S}, and produces corresponding visual edits (right). These edits can be viewed from an arbitrary camera viewpoint in a 3D-consistent manner. To this end, we propose a novel optimization that fine-tunes pre-trained latent diffusion models on multiple keyframes representing different viewpoints and time stamps (see Sec.~\ref{sec:FinetuneST}). Furthermore, we employ a new view- and time-aware Score Distillation Sampling (VT-SDS) that combines a pre-trained latent diffusion model with our fine-tuned model (see Sec.~\ref{sec:editing} and Eq.~\ref{equ: modified classifier-free}).} 
    \label{fig:my_label}
\end{figure*}

\label{sec:finetune}
\begin{figure}
    \centering    
    \includegraphics[width=0.5\textwidth]{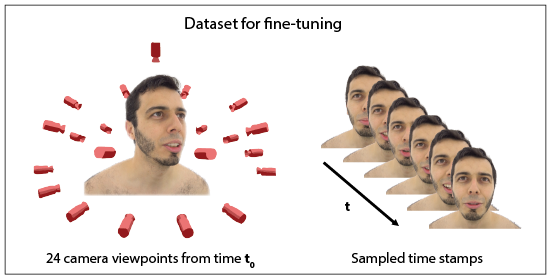}
    \caption{Sample viewpoints and time stamps used in our diffusion model fine-tuning (see Sec.~\ref{sec:optimization2})}. 
    \label{fig:fine-tuningdata}
\end{figure}

\section{Method}

Our objective is to edit dynamic 3D full human heads using a text prompt expressing the desired edit (see Fig.~\ref{fig:my_label}).
We assume the human heads are represented by dynamic NeRF-based models reconstructed with an existing method.
In this work, we use HQ3DAvatar~\cite{teotia2023hq3davatar} to obtain the dynamic human heads due to its high quality.
The editing is achieved by leveraging the prior knowledge of a large text-guided latent diffusion model (LDM)~\cite{StableDiff,rombach2021highresolution}.
We first describe preliminaries on HQ3DAvatar and LDM (Sec.~\ref{sec:preliminaries}).
We then introduce a new optimization strategy that adapts the LDM to capture the identity of the given head from different viewpoints and time stamps (Sec.~\ref{sec:finetune}).
This optimization step is essential for preserving the original head identity and details during editing, as we will show in experiments. 
Next, we discuss how we use the adapted LDM to edit the dynamic head with a text prompt (Sec.~\ref{sec:editing}).
Our approach enables personalized and targeted edits from textual inputs while maintaining the examined identity.
It also allows a wide range of text-driven edits of dynamic full heads and produces temporally consistent results.
For example, given a dynamic head and an exemplary textual input ``Turn her into a zombie'' or ``Make him look like Van Gogh'', our model can produce multi-view consistent NeRF edits that transfer the zombie or van Gogh styles to the target identity.
We evaluate our approach visually and numerically through a user study, and results show that we clearly outperform existing methods.
 
\subsection{Preliminaries} 
\label{sec:preliminaries}
\subsubsection{Neural Radiance Fields} 
The Neural Radiance Fields (NeRF) algorithm, as described in the paper by Mildenhall~\etal~\cite{mildenhall2021nerf}, employs a fully-connected deep neural network to represent a scene. The network is fed a continuous 5D coordinate that comprises a spatial location $\textbf{p}$ with coordinates $(p_x,p_y,p_z)$  
and a viewing direction $\textbf{v}$ $(v_\theta,v_\phi)$. The output of the network is the volume density $\sigma$ and the view-dependent emitted radiance at that location. By utilizing classic volume rendering techniques and querying 5D coordinates along camera rays, 
NeRF can project output colors and densities onto novel views to synthesize images.
The volume rendering process is differentiable, which allows for optimization of the representation using a set of input images with known camera parameters (extrinsic and intrinsic). These camera parameters are used to extract a per-pixel world-space ray parameterization that describes the 3D center $\mathbf{o}$ and direction $\mathbf{d}$ of the camera ray $\mathbf{r}(t) = \mathbf{o} + t\mathbf{d}$ corresponding to each pixel in each image. The expected color $C(\mathbf{r})$ of camera ray $\mathbf{r}(t)=\mathbf{o} + t\mathbf{d}$ with near and far bounds $\timenear$ and $\timefar$ can be calculated as follows:
\begin{align}
\label{eqn:volumerendering}
\Ctrue =& \int_{\timenear}^{\timefar}T(t)\absrp(\mathbf{r}(t))\mathbf{c}(\mathbf{r}(t),\mathbf{d})dt\,,
\\ \textrm{ where }
T(t) =& \expo{-\int_{\timenear}^{t}\absrp(\mathbf{r}(s))ds}. 
\end{align}

\subsubsection{Scene Representation}
\label{avatar}
To capture complete head performances, we make use of the dynamic volumetric representation of HQ3DAvatar \cite{teotia2023hq3davatar}, which learns a volumetric representation of the human head using multi-view RGB videos. 
We use this representation due to its high quality video results. While our method in principle could work with other dynamic NeRF models, examining this is outside the scope of this work.

HQ3DAvatar consists of two main stages. The first contains a deformation network $D$ that maps the input coordinates $\textbf{p}$ of the world space to deformed positions in the canonical space as follows
\begin{equation}
\label{eq:deformation}
   \textbf{p}_c = D(\textbf{p},\be) + \textbf{p} 
\end{equation}
Here, $\textbf{p}_c$ denotes the deformed coordinates in the canonical space, while $\be$ is a time embedding of the input RGB frames. This embedding is extracted using a pre-trained VGG-Face encoder~\cite{VGGFaceBMVC2015}.
The second stage contains an appearance network $A$ that predicts the radiance $\bc$ and volume density $\sigma$ for each deformed coordinate. This is written as: 
\begin{equation}\label{eq:apperancenet}
    A : (\textbf{p}_c, \bv, \be) \rightarrow (\bc, \sigma) \; \text{,}
\end{equation}
where $\bv$ is the viewing direction. 
To make the method computationally efficient, a multi-resolution hash-grid based representation is used along the lines of Mueller et al.~\cite{muller2022instant}. 
To this end, the appearance network $A$ consists of two main parts. The first is the multi-resolution hash grid, while the second is an MLP-based network that outputs the radiance $\bc$ and volume density $\sigma$. This MLP is conditioned on the time embedding $\be$ as well as the viewing direction $\bd$.
Once we have the radiance field representation of the scene, we
use standard volumetric integration to synthesize color \textbf{C} for each
ray ${\bm r}(t)$ using Eqn.~\eqref{eqn:volumerendering}.
For more details on the method including the network design and the efficient rendering, please refer to the manuscript of \cite{teotia2023hq3davatar}. 

\subsubsection{Latent Diffusion Models} 
Latent diffusion models (LDMs)~\cite{rombach2021highresolution} are a class of Denoising Diffusion Probabilistic Models (DDPMs)~\cite{ho2020denoising} that use an vector-quantized auto-encoder~\cite{van2017neural} to translate an input image into a latent space in which a text-conditioned DDPM is trained.
The encoder $\mathcal{E}$ processes a given image $\mathcal{I} \in \mathbb{R}^{H \times W \times 3}$ to a latent representation $\mathbf{z}$, such that $\mathbf{z}=\mathcal{E}(\mathcal{I})$. The decoder $\mathcal{D}$ reconstructs the estimated image $\tilde{\mathcal{I}}$ from the latent, such that $\tilde{\mathcal{I}}=\mathcal{D}(\mathbf{z})$ and $\tilde{\mathcal{I}} \approx \mathcal{I}$.
The diffusion model is trained to generate images in the latent space of the encoder.
Similar to other types of generative
models~\cite{mirza2014conditional}, diffusion models
are in principle capable of modeling conditional distributions of the form $p(x \vert s)$, where $s$ is the conditioning variable.
Conditional latent diffusion model are learned to optimize the following loss:
\begin{equation}
\small
\mathbb{E}_{\mathcal{E}(\mathcal{I}), s, \boldsymbol{\epsilon} \sim \mathcal{N}(0,1), t}\left[\left\|\boldsymbol{\epsilon}-\boldsymbol{\epsilon}_\theta\left(\mathbf{z}_t, t, s\right)\right\|_2^2\right],
\label{equ: ldm}
\end{equation}
where $t$ is the diffusion time step and $\mathbf{z}_t$ is the noisy latent code at time $t$. $\boldsymbol{\epsilon}$ is the noise sample, $\boldsymbol{\epsilon}_\theta$ is the denoising model with parameters $\theta$ and $s$ is the conditioning input.
During training, $\boldsymbol{\epsilon}_\theta$ is optimized.
At inference, a latent code is generated by randomly sampling a noise tensor and denoising it iteratively based on a conditioning input.

\subsection{Fine-Tuning Text-to-Image Latent Diffusion Model}
\label{sec:FinetuneST}

A key challenge in dynamic full head editing is to preserve the original characteristics of the head (\eg, its identity, details, motions, \etc) rather than creating a completely different head.
Editing using the original LDM would soon lead to drifts of these characteristics due to information leakage.
A potential way to alleviate this problem is to fine-tune the LDM on images of the given head using DreamBooth~\cite{ruiz2022dreambooth}.
However, unlike DreamBooth which aims to sample new 2D images and thus only needs to capture the object identity, we want to edit a 3D head across different viewpoints and time stamps.
Thus, ideally, the LDM should not only be identity-aware but also viewpoint-aware and time-aware.
Our investigations revealed that implementing DreamBooth for multiple concepts (\ie, multiple viewpoints and time stamps) is also prone to concept leakages, producing suboptimal editing.
Hence, below we detail our new optimization strategy which is designed specifically for associating multiple concepts for different viewpoints and time stamps.

\subsubsection{Optimization.}
\label{sec:optimization2}
 In our work, we fine-tune the LDM of Stable Diffsion~\cite{StableDiff,rombach2021highresolution}. To this end, we use images of the given dynamic 3D head from different viewpoints and time stamps (see Fig.~\ref{fig:fine-tuningdata}), denoted as $\{\bx_i; i \in \{1,...,n\}\}$.
 How we select the images will be discussed in Sec.~\ref{sec:sampling}.
 We then assign a label$\bP_{i}$ to each of these images, using the format `photo of a [identifier] [class noun]'.
 The identifier is a one-of-a-kind code of 10 characters, unique for each image, while the class noun is `man' or `woman' depending on the gender of the given head. Each identifier is initialized via a random word generator.Our aim is to fine-tune a pretrained text-to-image LDM $\bzeta_\theta$ so that, given an initial noise map $\bepsilon \sim \mathcal{N}(\bzero, \bI)$ and a conditioning vector $\mathbf{s}_{i}=\Gamma(\bP_{i})$ produced using a text encoder $\Gamma$, the fine-tuned LDM $\hat\bzeta_\theta$ will reconstruct the image $\bx_i$. The model is fine-tuned using a squared error loss to denoise a variably-noised image or latent code $\bz_{t,i} \coloneqq \alpha_t \mathcal{E}(\bx_i) + \beta_t \bepsilon$ as follows:
\begin{equation}
\label{eq:diffusion}
    \Eb{\bx_i,\mathbf{s}_{i},\bepsilon,t}{w_t \|\bzeta_\theta(\alpha_t \bx_i + \beta_t \bepsilon, \mathbf{s}_{i}) - \mathcal{E}(\bx_i) \|^2_2}
\end{equation}

where $\alpha_t, \beta_t, w_t$ are terms that control the noise schedule and sample quality. These terms are a function of the diffusion process time $t \sim \mathcal{U}([0, 1])$~\cite{ruiz2022dreambooth}. To overcome potential language drift of language models~\cite{lee2019countering,lu2020countering}, we incorporate Class-specific Prior Preservation Loss~\cite{ruiz2022dreambooth}. 
In practice, we use a batch size of 3 during fine-tuning, which corresponds to 3 randomly sampled $\bx_i$ with different viewpoints or time stamps. We observe that using different noise $\bepsilon$ within each batch may lead to concept leakage between different identifiers.
Thus, we use a shared noise $\bepsilon$ within each batch, which helps the identifiers to capture the variations in the image avoiding any leakage.

\subsubsection{Time Embedding Sampling.}
\label{sec:sampling}

Here we introduce how we select the images $\{\bx_i\}$ used to fine-tune the LDM.
As discussed above, $\{\bx_i\}$ should include multiple viewpoints and time stamps.
Thus, we use all camera views of the first frame as the multiview images.
In our experiments we use a multiview camera rig equipped with 24 RGB cameras around the head (see Sec.~\ref{sec:DataCapture}). Hence, we use all these 24 camera viewpoints.
We also include 6 other frames from the frontal camera view, which should be as diverse as possible.

To achieve this, we employ an empirical strategy. We generate embeddings $\textbf{e}_{j}$ in HQ3DAvatar (see Eq.~\eqref{eq:deformation})
 
for each of the $m$ frames in a given dynamic head. 
We calculate the mean of these $m$ embeddings and then select 6 embeddings and their corresponding frames that exhibit the greatest variation from the mean based on their absolute difference. We make sure to discard similar neighbouring frames with similar deformations to avoid redundancy. 
\subsection{Text-guided Dynamic NeRF Editing}
\label{sec:editing}
With a pre-trained HQ3DAvatar on a specific identity, we perform text-driven editing by optimizing the appearance network $A$ (see Eq.~\ref{eq:apperancenet}), while keeping the deformation network $D$ fixed. 
In other words, we edit the appearance in HQ3DAvatar's canonical space. Please refer to Fig.~\ref{fig:my_label} for an overview of this editing process. 
Our text-driven editing is developed based on Score Distillation Sampling (SDS) loss~\cite{poole2022dreamfusion}, which supervises an image to follow the text prompt using a text-to-image LDM.
Here, we render an image $\bx$ at each optimization step by randomly sampling from the camera viewpoints and time stamps corresponding to $\{\bx_i\}$ (Sec.~\ref{sec:sampling}). 

At every step of the optimization process, a random diffusion time instant $t$ 
is sampled and noise is injected into the rendered image $\bx$: 
\begin{equation}
    \bx_t = \bx + \bepsilon_t,
\end{equation}
where $\bepsilon_t$ is the noise map generated via a noising function $Q(t)$.

With our modified Score Distillation Sampling loss (see Sec.~\ref{modified-sds}), the gradients for score distillation are calculated on a per-pixel basis as follows:
\begin{equation}
  \nabla_x\mathcal{L}_{SDS} = w(t)\left( \bepsilon_t-{\Psi}(\bx_t, t, \mathbf{s}, \mathbf{s}_{i}) \right).
\end{equation}
Here, $w(t)$ is a weighting function following~\cite{poole2022dreamfusion}, $\mathbf{s}$ is the text embedding of the user-input text for editing the dynamic NeRF, 
and $\mathbf{s}_{i}$ is the same as in Eq.~\ref{eq:diffusion}.
Furthermore, ${\Psi}(\bx_t, t, \mathbf{s}, \mathbf{s}_{i})$ is the noise predicted by a combination of our fine-tuned 
and a pre-trained diffusion model  
 given $\bx_t$, $t$, $\mathbf{s}$ and $\mathbf{s}_{i}$ as we will introduce later (Eq.~\ref{equ: modified classifier-free}).

Moreover, we utilize a regularizer for the density field generated by A~\cite{melaskyriazi2023realfusion} as follows: 
\begin{equation}
\label{sigmareg}
    \mathcal{L}_{\text{entropy}} = \omega \cdot \log_2 (\omega) - (1-\omega) \cdot \log_2 (1-\omega). 
\end{equation}
Here, $\omega$ is the cumulative sum of density weights computed along each ray in the scene. This regularizer is an entropy loss that promotes the points to be either completely transparent or completely opaque. 

\subsubsection{Modified Score Distillation Sampling}
\label{modified-sds}
Inspired by recent work \cite{zhang2022sine} on reducing overfitting and severe language drift in fine-tuned text-driven diffusion models,
we utilize our fine-tuned model to provide content features. These features are combined with scores
from the pre-trained model in a manner similar to classifier-free guidance~\cite{ho2022classifier}.

To this end, we use the notation $\boldsymbol{\hat{\zeta}}_\theta$ to refer to the fine-tuned denoising model, and $\boldsymbol{\zeta}_\theta$ to refer to the pre-trained text-to-image model. During Score Distillation Sampling (SDS), we guide the pre-trained model with our fine-tuned model by using a linear combination of the noise estimated from each model, for a specified range of optimization steps.
Thus, the noise estimation in a SDS step  
can be defined as:
\begin{equation}
\small
\begin{aligned}
{\boldsymbol{\Psi}}\left(\mathbf{x}_t, t, \mathbf{s}, \mathbf{s}_{i}\right)=& w\left(v \boldsymbol{\zeta}_\theta\left(\mathbf{x}_t, \mathbf{s}\right)+(1-v) \hat{\boldsymbol{\zeta}}_\theta\left(\mathbf{x}_t, {\mathbf{s}_i}\right)\right) \\
&+(1-w) \boldsymbol{\zeta}_\theta\left(\mathbf{x}_t\right),
\end{aligned}
\label{equ: modified classifier-free}
\end{equation}
where $w$ is the overall guidance weight and $v$ stands for the model guidance weight, which depends on $t$ as we will discuss later. 
$\hat{\mathbf{s}}_i$ is the same as in Eq.~\ref{eq:diffusion} and $\mathbf{s}$ is the target language conditioning obtained from the target prompt (see Sec.~\eqref{sec:optimization2}). 

To ensure that the edited dynamic neural radiance field remains faithful and free from overfitting artifacts, we use Eq.~\eqref{equ: modified classifier-free} with $0.5\leq v \leq 0.7$ for sampling when $t>K$, and $v=1$ for sampling when $t \leq K$. Unless stated otherwise, we use $K=600$. We follow~~\cite{lin2022magic3d} and use an annealed SDS loss function that gradually lowers the maximum time-step used to sample $t$. 
This enables SDS to emphasize high-frequency information once the edit's outline has been established.
In the ablation study of Sec.~\ref{Sec:Ablation}, we show that using Eq.~\eqref{equ: modified classifier-free} for Score Distillation Sampling is essential for generating good edits in both the the spatial and temporal domains. 

\section{Experiments}

In this section, we evaluate the performance of our method subjectively and numerically. 

\textbf{Performance measures.} We asses three main aspects of the generated results. First, their ability to respect the target text prompt. Here, it is important to maintain the integrity of the input identity. Second, we asses the ability of generating edits that are 3D-consistent, and thus can be rendered from an arbitrary camera viewpoint. Third, we assess the temporal coherency of the generated edits. Visual results are shown throughout figures and the supplemental video. Numerical results are extracted from a user study. 
Results show that our method is capable of producing a wide variety of text-driven edits that are 3D-consistent and temporally coherent. 

\textbf{Baselines.} We compare against two text-driven image synthesis baselines: Dream Fields~\cite{jain2021dreamfields} and Instruct-NeRF2NeRF~\cite{instructnerf2023}.
We also compare against an implementation that combines the diffusion model fine-tuning method of DreamBooth~\cite{ruiz2022dreambooth} together with the 3D text-based editing method of DreamFusion~\cite{poole2022dreamfusion}. 
Results show that our method clearly outperforms the state of the art visually and numerically. 

In the next two sections we discuss implementation details and the multi-view data that is used in our experiments.
Subsequently, we discuss the user study that we perform in assessing our experiments (Sec.~\ref{sec:UserStudy}).
We show in Sec.~\ref{sec:exp:editing} extensive evaluation of our method using varying prompts. 
In Sec.~\ref{sec:SOTAComparison}, we compare against related methods subjectively and objectively. 
Finally, we investigate the various design choices of our method in an ablation study (Sec.~\ref{Sec:Ablation}). Here we investigate our two main contributions. 
First, we investigate the importance of our fine-tuning strategy (Sec.~\ref{sec:FinetuneST}) which incorporates multiple camera viewpoints and different time stamps.  
We also investigate the importance of incorporating Eq.~\ref{equ: modified classifier-free} in the Score Distillation Sampling as discussed in Sec.~\ref{sec:editing}. 
Results show that all our design choices contribute positively to the final output. 

\subsection{Implementation Details}
We employ a consistent set of parameters for all experiments, without optimizing them specifically for each scenario. We utilize the open-source Stable Diffusion model \cite{rombach2021highresolution} as our prior for the diffusion model. This model was trained on the LAION dataset \cite{schuhmann2022laion}, which consists of pairs of text and images. We render our images at a resolution of 256px. Since the Stable Diffusion model is specifically designed for images with a resolution of 512px, we first upsample our renders to 512px before passing them to the Stable Diffusion's latent space encoder, which is a Variational Autoencoder (VAE). We use classifier-free guidance of strength 10. Additionally, we set the VT SDS ratio to v=0.6 and K=600.
We optimize the head avatar for a single prompt using the Adam~\cite{kingma2014adam} optimizer with learning rate 1\text{e-}3 for 10000 iterations.
The optimization process takes approximately 60 minutes on a single A100 GPU. For our fine-tuning step, we optimize the diffusion model for a total of 28000 steps using the Adam optimizer~\cite{kingma2014adam} with image size 512px, batch size 3 and a learning rate 5\text{e-}5.

\subsection{Data Capture} 
\label{sec:DataCapture}
\begin{figure*}
    \centering    \includegraphics[width=\textwidth]{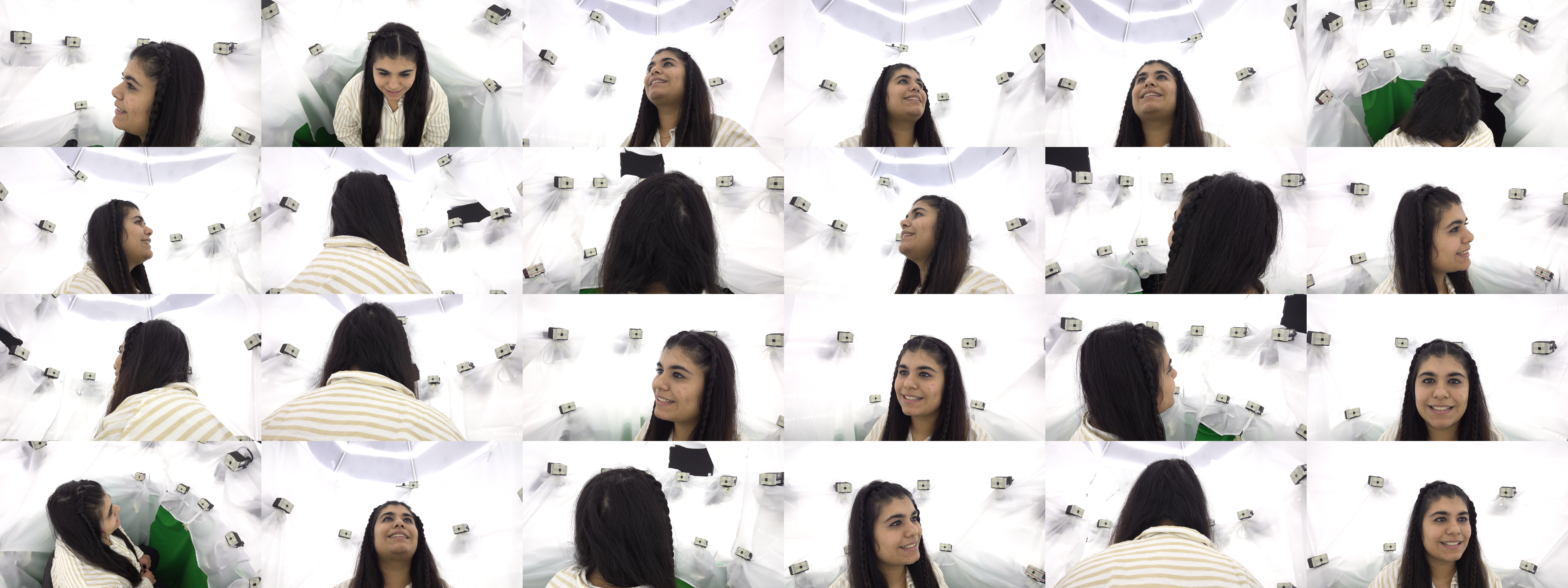}
    \caption{We trained our method using data captured from a multi-view video camera rig. The rig contains 24 video cameras positioned around the head. The figure shows an example capture from each camera viewpoint}
    \label{fig:TanyaCollage}
\end{figure*}

Our method is trained on multi-view data. We use a 360-degree camera rig equipped with 24 Sony RXO II cameras that are hardware-synced and capable of recording 25 frames per second at a 4K resolution. These cameras are positioned in a way to capture the entire human head, including the scalp's hair. They are also accompanied by LED strips to provide uniform illumination. 
The cameras are calibrated using a static structure with distinctive features. The intrinsic and extrinsic parameters are estimated using Metashape~\cite{agisoft2020metashape}. Background subtraction is carried out using the matting approach of Lin~\etal~\cite{lin2021real} to eliminate static elements like wires, cameras, and other objects. To simplify the process of background subtraction, a diffused white sheet is placed inside the rig, which contains holes for each camera lens. 
Please refer to Fig.~\ref{fig:TanyaCollage} for an overview of data captured and used in our experiments. 
We use data collected by this camera rig for all our experiments, including evaluating related methods.
We show results on 5 identities performing a variety of expressions. Some identities are also reading a set of sentences known as Pangrams~\footnote{https://callibeth.com/downloads/pangrams111.pdf}. 
Here, each sentence contains all the 26 Latin letters.
Our videos are recorded at 25 frames per second, and are between 300-500 frames long.   

\subsection{User Study}
\label{sec:UserStudy}

It is challenging to evaluate text-driven visual edits numerically due to the absence of ground-truth data. In fact, one text prompt could have several different possible visual edits. 
Current methods used two main strategies for numerical evaluations. One approach is to measure the alignment of the produced edits with the input text prompt using the CLIP space~\cite{jain2021dreamfields,poole2022dreamfusion,instructnerf2023}. This method, however, does not evaluate the temporal coherency of the solution. In addition, it is expected to favour methods that optimize their solution in the CLIP space. Another strategy for numerical evaluation is to perform a user study, as adapted by~\cite{wang2022nerf,lin2022magic3d}.
We believe this strategy is more suitable as it is not tied to a specific text-image embedding, and due to the subjective nature of the examined problem.

Motivated by this, we designed a user study that assess several important aspects of text-driven video edits. 
For a given identity and a given text prompt, our user study shows four videos side-by-side. 
The first is the original input as produced by HQ3DAvatar~\cite{teotia2023hq3davatar}, while the remaining videos are the output of three different text-driven editing methods. 
The order of these three videos is randomly shuffled. 
Each video is around 19 seconds, featuring either a talking person or different facial expressions captured by a rotating camera.
Participants were asked to watch the video and were given the option to replay it as desired. 
They were then asked to answer the following set of questions.
\begin{itemize}
    \item  Q1: Which method better retains the identity of the input sequence (identity preservation)?
    \item  Q2: Which method better follows the given input textual prompt (prompt preservation)?
    \item  Q3: Which method better maintains temporal consistency (temporal consistency)?
    \item  Q4: Which method is better overall considering the above 3 aspects (identity preservation, textual preservation, temporal consistency)?
\end{itemize}
As shown, the first three questions are designed to assess (from the top) the identity preservation, the prompt preservation and the temporal consistency of the output. 
The user gives an answer to each of these three questions, and hence a specific method output could be chosen as the best one in just a subset of these questions. 
Finally, the fourth question is a measure of the overall quality considering all these three aspects (identity preservation, textual preservation, temporal consistency). 

\subsection{Text-driven Full Head Editing}
\label{sec:exp:editing}

\begin{figure*}
    \centering    \includegraphics[width=\textwidth]{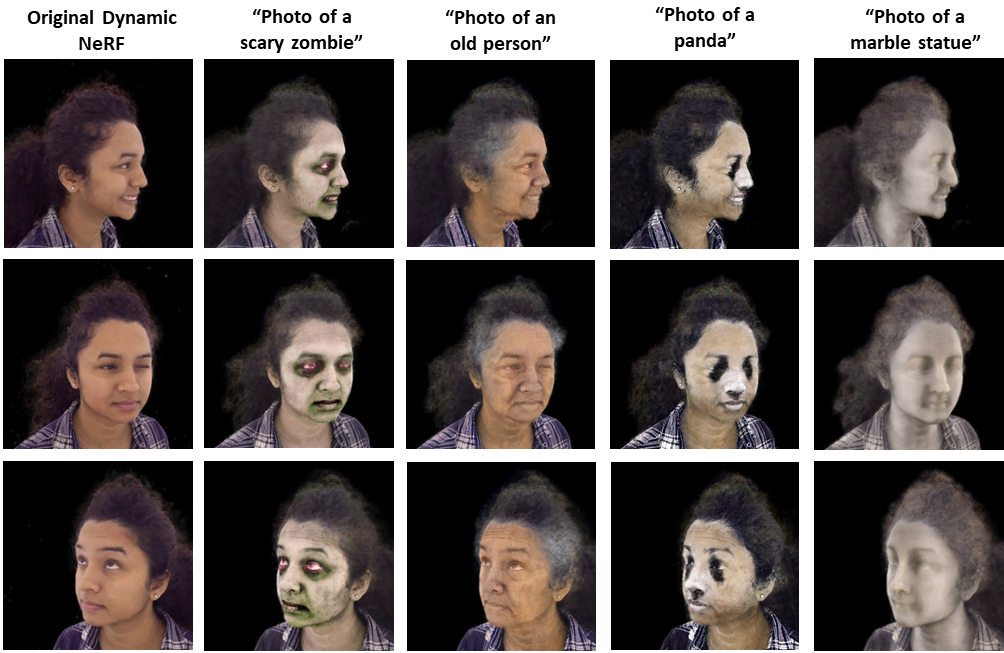}
    \caption{Our method produces compelling text-driven visual edits for different text prompts. Note the good sharp edits in the eyes (see the second column) and how the fourth column 
    show edits in the facial geometry. Our results are 3D- and temporally consistent as can be seen in the supplemental video.}
    \label{fig:Navami_results}
\end{figure*}
\begin{figure*}
    \centering    \includegraphics[width=\textwidth]{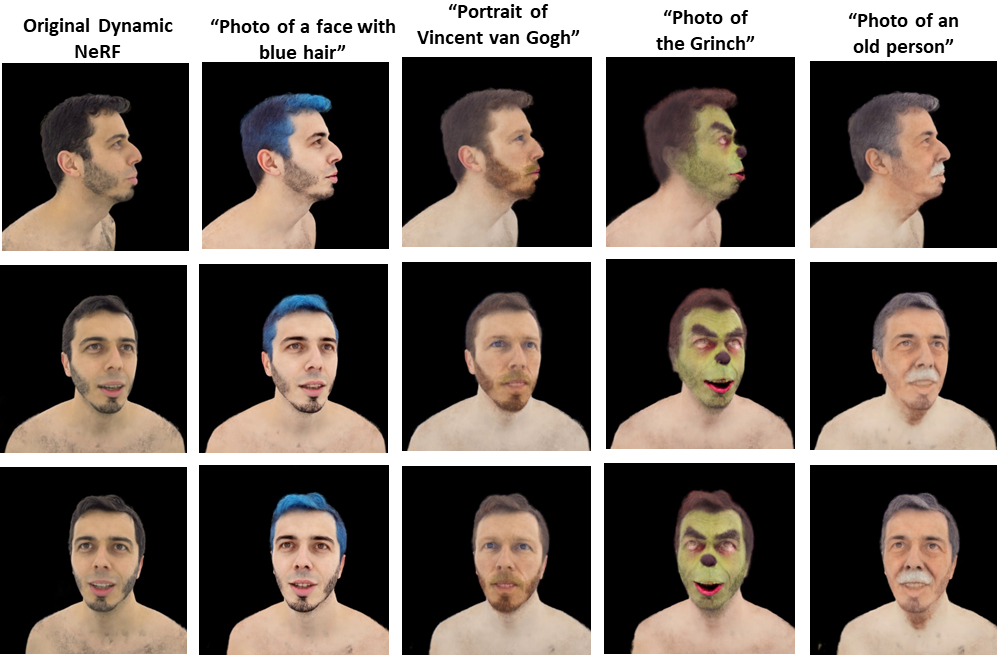}
    \caption{Our method produces pleasing text-driven visual edits for different prompts. Note for instance the geometrical edits in the fourth column and how our method can handle interesting 
    prompts such as "photo of an old person" as shown in the last column. Our method can also edit specific regions as instructed by the prompt (see blue hair, second column).
    Our results are 3D- and temporally consistent and maintains the original identity. 
    } 
    \label{fig:Diogo_results}
\end{figure*}
\begin{figure*}
    \centering    \includegraphics[width=\textwidth]{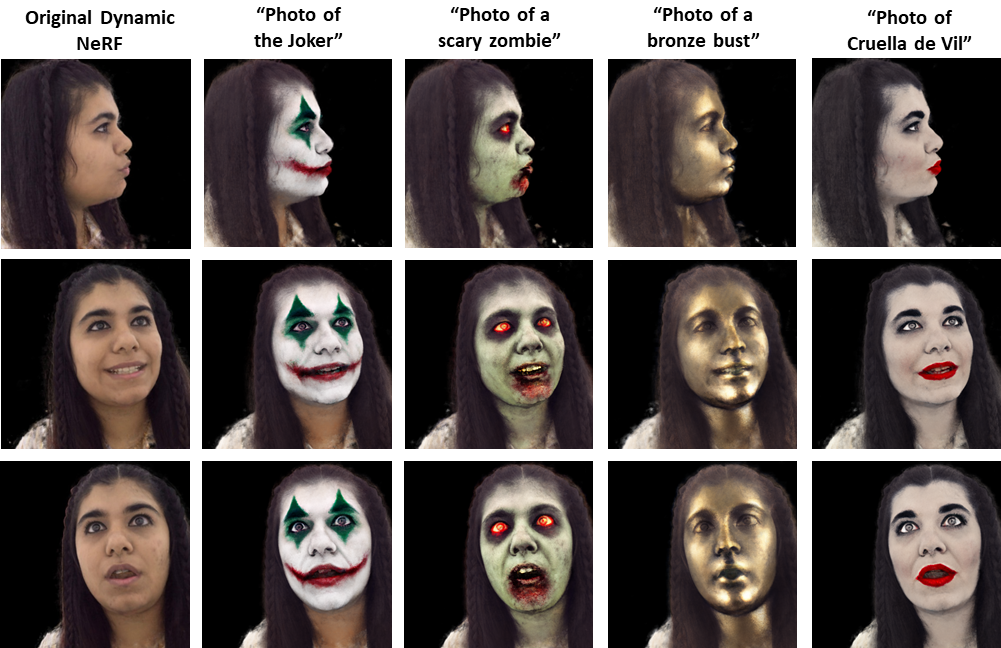}
    \caption{Our method produces compelling text-driven visual edits for different text prompts. 
    Note how the bronze bust editing prompt introduces the effects specific to metals and makes the facial texture more uniform. Note the sharp transition between the edited red eyes and the skin in the third column. All results preserve the original identity and are 3D- and temporally consistent, which can be observed in our supplemental video.
    } 
    \label{fig:Tanya_results} 
\end{figure*}
\begin{figure*}
    \centering    \includegraphics[width=\textwidth]{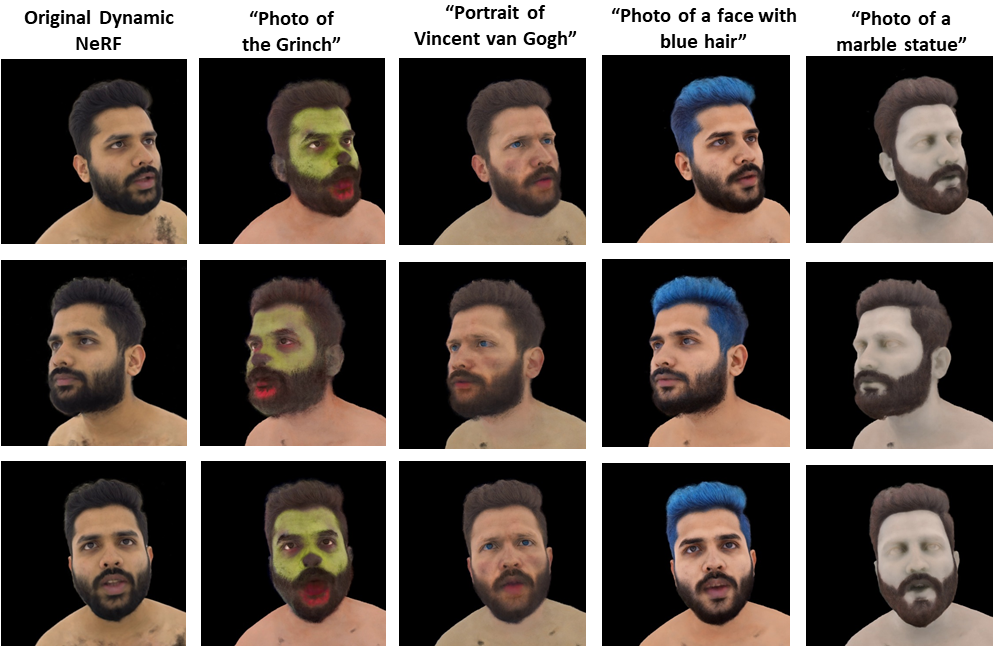}
    \caption{Our method produces compelling text-driven visual edits for different text prompts. Our results maintains the original identity and are 3D- and temporally consistent as shown in the supplemental video. 
    } 
    \label{fig:Basavaraj_results}
\end{figure*}
\begin{figure*}
    \centering    \includegraphics[width=\textwidth]{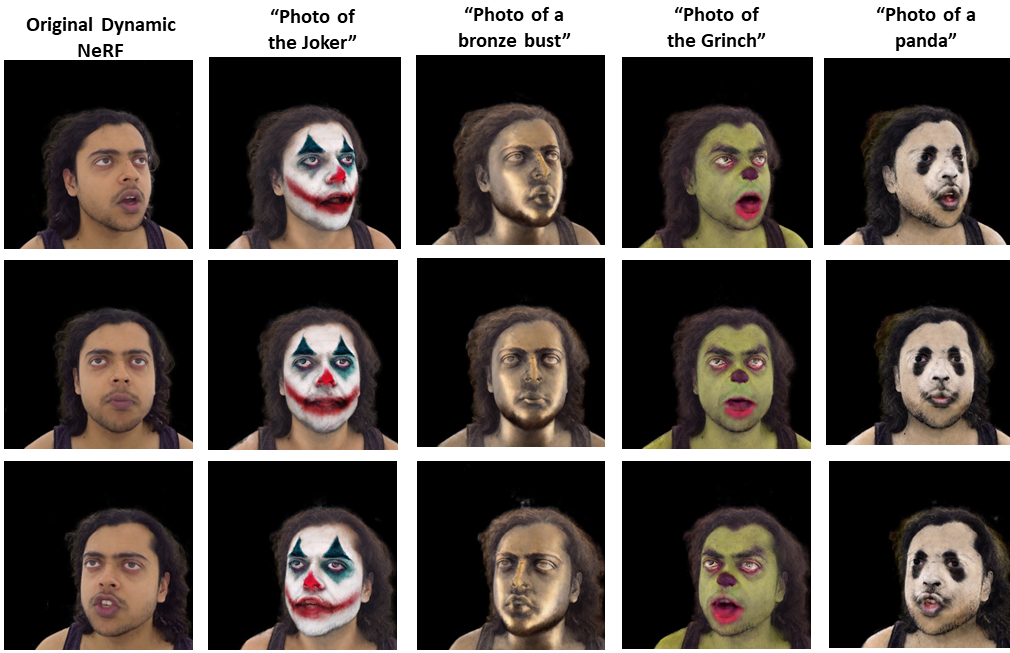}
    \caption{Our method produces compelling text-driven visual edits for different text prompts. Note how our method can change the appearance (all columns) and the geometry (last two columns).
    Our results are 3D- and temporally consistent.  
    } 
    \label{fig:Kartik_results}
\end{figure*}

Fig.~\ref{fig:Navami_results}-\ref{fig:Basavaraj_results} show various edits generated by our method. Here, we show results for several identities with various input text prompts. 
Subjects are talking and performing various expressions while we show results from different camera viewpoints.
Results show that our method can handle a wide variety of text-driven visual edits. This includes both photorealistic (e.g. Fig.~\ref{fig:Navami_results} and Fig.~\ref{fig:Diogo_results} "Photo of an old person"), and non-photorealistic (Fig.~\ref{fig:Navami_results} and Fig.~\ref{fig:Kartik_results} "Photo of a Panda") edits. We can edit specific regions of the face according to the input text prompt. 
For instance, Fig~\ref{fig:Diogo_results} (second column) and Fig.~\ref{fig:Basavaraj_results} (fourth column) show that we can edit the color of hair in isolation, which respects the text prompt of "Photo of a face with blue hair". 
Our method can also edit the geometry so that it is inline with the text prompt. For instance, see "Photo of the Grinch" (Fig.~\ref{fig:Diogo_results} and Fig.~\ref{fig:Kartik_results}) and "Photo of panda" (Fig.~\ref{fig:Kartik_results}).
Our method handles a variety of facial expressions and head movements. 
This includes normal speech (Fig.~\ref{fig:Diogo_results}), smiling (Fig.~\ref{fig:Tanya_results}, middle row), and extreme expressions (Fig.~\ref{fig:Tanya_results}, last row). 
Results also show that our method produces edits that are 3D consistent as well as temporally coherent. This is best observed in the supplemental video.
Last but not least, our method achieves this wide range of edits while maintaining the integrity of the input identity. Please refer to the supplemental video for more results. 

\subsection{Comparison Against Related Methods}\label{sec:SOTAComparison}
We compare against two recent 3D text-based editing methods. That is InstructNeRF2NeRF~\cite{instructnerf2023} and Dream Fields~\cite{jain2021dreamfields}.
In both methods, we edit only the reconstructed HQ3DAvatar's
appearance in the canonical space, while keeping the deformation network fixed. We then use the deformation network to warp the edited appearance to the remaining time stamps. 
Since the original Dream Fields method~\cite{jain2021dreamfields} generates the entire image purely from text, we implemented a version which is initialized by our reconstructed HQ3DAvatar model. 
This is done by optimizing the HQ3DAvatar's appearance via the CLIP loss~\cite{CLIP}.
We call this implementation Dream Fields++ in the rest of the paper. We also call InstructNeRF2NeRF in our figures InstructN2N for brevity. 

\begin{figure*}
    \centering    \includegraphics[width=\textwidth]{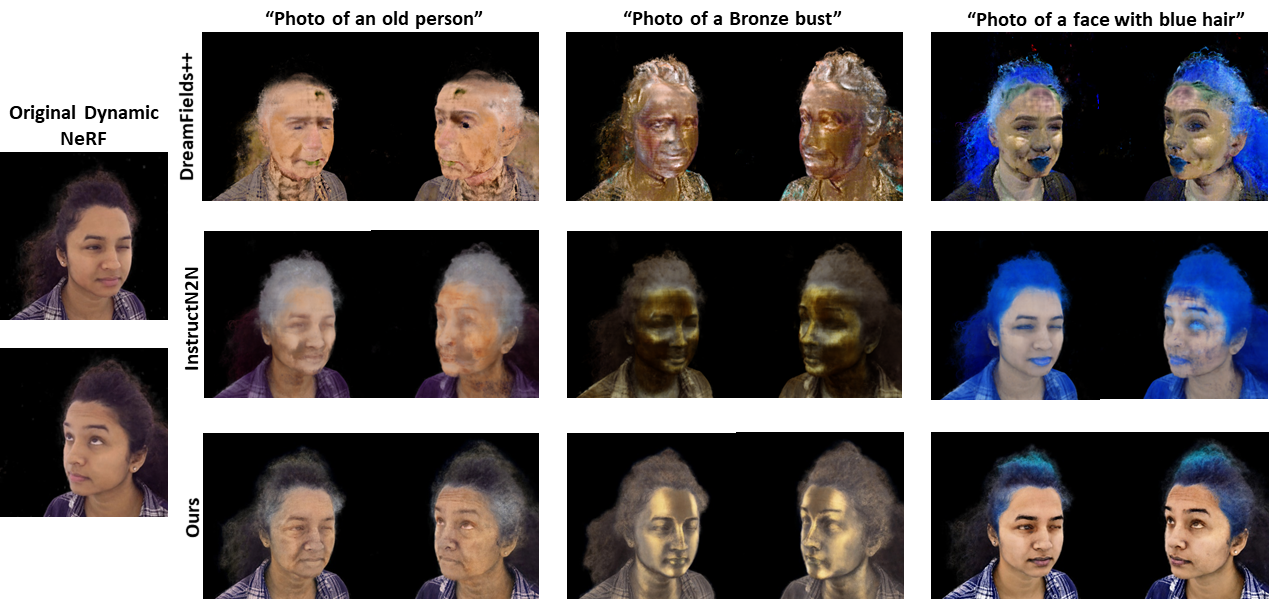}
    \caption{Our approach outperforms Dream Fields++~\cite{jain2021dreamfields} and InstructeNeRF2NeRF~\cite{instructnerf2023} spatially and temporally. It maintains the original identity and generates edits that are inline with the target prompts. Please see the supplemental video and the user study (Tab.~\ref{tab:userstudy}) for video coherency.} 
    \label{fig:Navami_comparisons1}
\end{figure*}

\begin{figure*}
    \centering    \includegraphics[width=\textwidth]{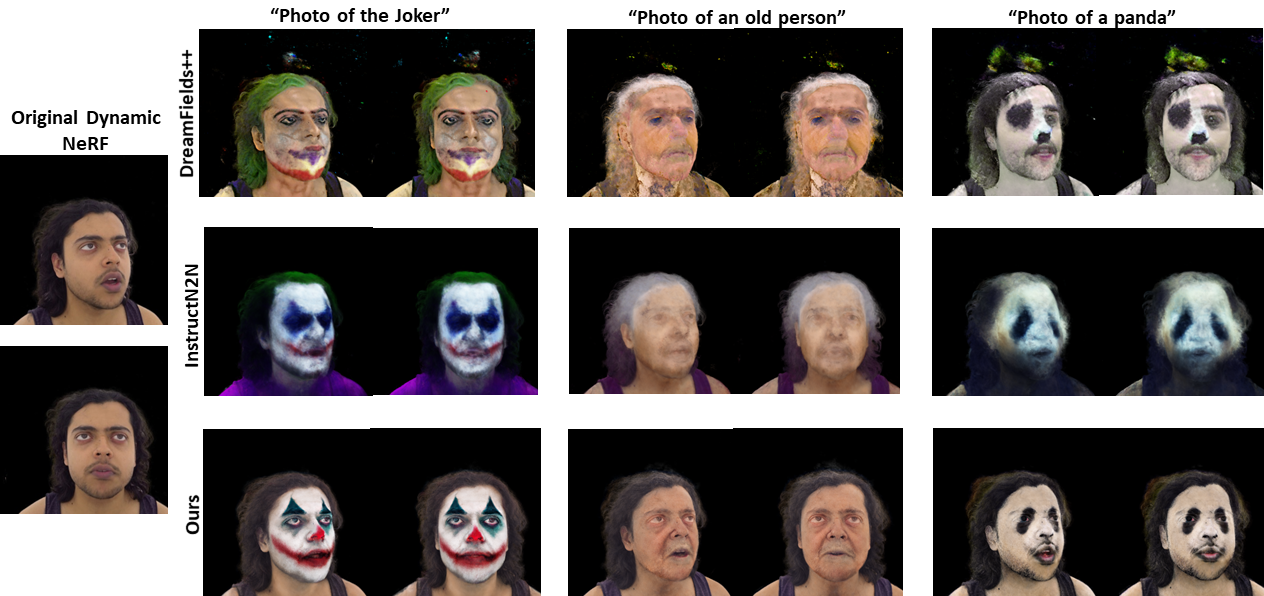}
    \caption{Our approach outperforms Dream Fields++~\cite{jain2021dreamfields} and InstructeNeRF2NeRF~\cite{instructnerf2023} spatially and temporally. It maintains the original identity and generates edits that are inline with the target prompts. Please see the supplemental video and the user study (Tab.~\ref{tab:userstudy}) for video coherency.} 
    \label{fig:kartik_comparison}
\end{figure*}

Fig.~\ref{fig:Navami_comparisons1}-\ref{fig:kartik_comparison} show results of different methods for different text prompts. Results show that Dream Fields++ generates significant artifacts that destroys the integrity of the input image. These edits are also clearly not inline with the target text prompt. Furthermore, Fig.~\ref{fig:Navami_comparisons1} (third column) shows that Dream Fields++ generates edits in unwanted regions. Here, despite the text prompt says "Photo of a face with blue hair", Dream Fields++ turns the lips blue. 
Similarly, InstructNeRF2NeRF generates edits that are not well aligned with the target text prompts. For instance, see Fig.~\ref{fig:Navami_comparisons1} and Fig.~\ref{fig:kartik_comparison} with prompt "Photo of an old person". InstructNeRF2NeRF could also destroy the original identity (see Fig.~\ref{fig:kartik_comparison}, last column) and some times it generates edits in wrong regions (see Fig.~\ref{fig:Navami_comparisons1}, last column, lips). 
In contrast, our method generates edits that are more inline with the target text prompt. In addition, it maintains the integrity of the input image and produces clearly more temporally consistent results. For this, please refer to the supplemental video and the user study (discussed next).

\begin{table*}[]
\begin{tabular}{|l|l|l|l|l}
\cline{1-4}
                           & Dream Fields++ & InstructNeRF2NeRF & StudioAvatar &  \\ \cline{1-4}
Q1: Identity preservation & 4.6            & 7.2               & 88.2         &  \\ \cline{1-4}
Q2: Prompt preservation    & 4.4            & 22.2              & 74.4         &  \\ \cline{1-4}
Q3: Temporal consistency   & 6              & 8.3               & 85.6         &  \\ \cline{1-4}
Q4: Overall                & 3.9            & 10.6              & 85.4         &  \\ \cline{1-4}
\end{tabular}
\caption{Reporting the results of our user study which included responses from 48 participants. 
Each participant was presented with the outputs of different methods and was asked to pick his/her preference with respect to the four questions listed in Sec.~\ref{sec:UserStudy}.
The table reports the percentages at which a method was rated the best with respect to a specific question. Hence the sum of each row should add up to 100.
Our method was rated the best in overall quality (Q4) $85.4\%$ of the time. This compares favorably to $3.9\%$ and $10.6\%$ for Dream Fields++ and InstructeNeRF2NeRF respectively. 
Our method was also clearly rated the best in the remaining questions.}\label{tab:userstudy}
\end{table*}

\begin{figure*}
    \centering    \includegraphics[width=\textwidth]{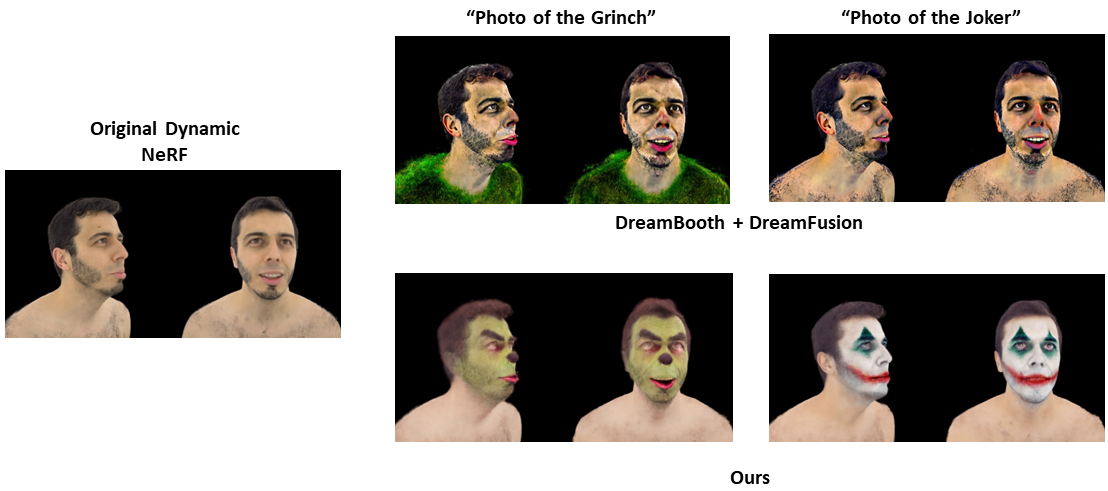}
    \caption{An implementation that combines DreamBooth~\cite{ruiz2022dreambooth} with DreamFusion~\cite{poole2022dreamfusion} has limited edibility (see top row). This is contrast to our method which produces edits that are clearly more inline with the target prompts.} 
    \label{fig:diogo_dreambooth}
\end{figure*}

Following Sec.~\ref{sec:UserStudy}, we perform a user study to compare our method against Dream Fields++ and InstructNeRF2NeRF. Here, we examine a total of 3 identities, each processed with 3 different prompts. Thus, the users were presented with a total of 9 videos, each consisting of four videos playing side-by-side as discussed in Sec.~\ref{sec:UserStudy}. Each time, the users were asked to asses various spatial and temporal aspects of the different methods by answering the four questions listed in Sec.~\ref{sec:UserStudy}. 
Hence, in total they were asked to answer these questions 9 times, where in each time the order of different methods was randomly shuffled.
Tab.~\ref{tab:userstudy} summarizes the findings of this user study. In total, 48 users participated in the study. Our technique is rated clearly the best in all questions. For instance, users rated our method the best in the overall quality $85.4\%$ of the time. 
This compares favorably to $3.9\%$ and $10.6\%$ for Dream Fields++ and InstructeNeRF2NeRF respectively.
One can create another straightforward baseline by combining the fine-tuning approach of DreamBooth~\cite{ruiz2022dreambooth} together with the 3D text-based editing method of DreamFusion~\cite{poole2022dreamfusion}.
We noticed, however, that this approach suffers from overfitting and usually gives poor edits that do not follow the prompt well, as shown in Fig.~\ref{fig:diogo_dreambooth}. Here, it is quite clear that our method is clearly superior. 

\subsection{Ablation Study}
\label{Sec:Ablation}

\begin{figure}
    \centering    \includegraphics[width=0.5\textwidth]{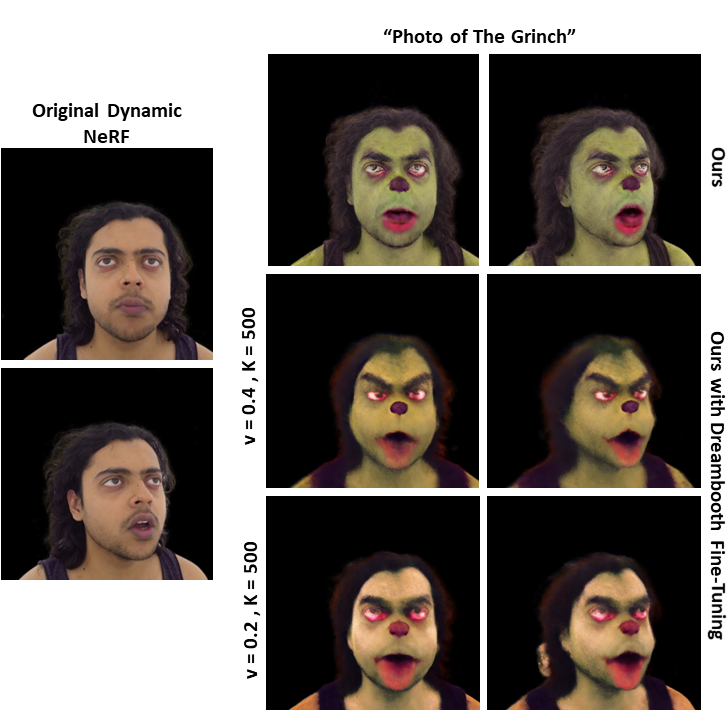}
    \caption{Comparison between our viewpoint-and-time-specific fine-tuning (Sec.~\ref{sec:optimization2}) and DreamBooth fine-tuning. Ours has fewer artifacts and better preserves the identity, showing the importance of viewpoint-and-time-specific fine-tuning. } 
    \label{fig:importance of optimization}
\end{figure}

\begin{figure}
    \centering    \includegraphics[width=0.5\textwidth]{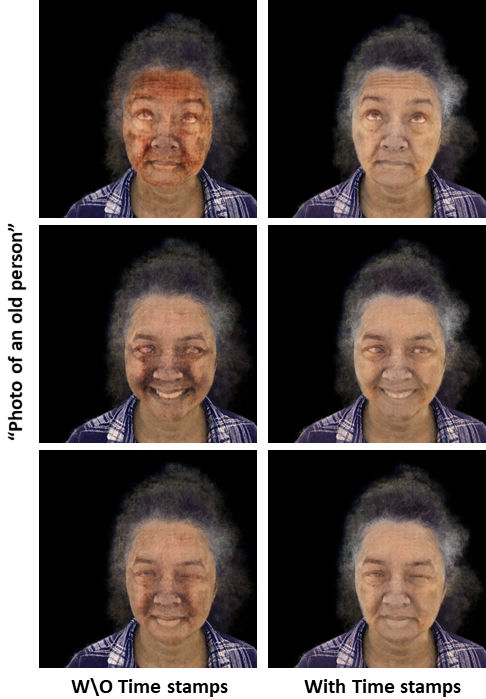}
    \caption{Fine-tuning LDM (Sec.~\ref{sec:FinetuneST}) without any temporal information leads to clear temporal inconsistencies. Notice the sharp temporal transition in the overall color.} 
    \label{fig:Navami_time}
\end{figure}

\begin{figure}
    \centering    \includegraphics[width=0.5\textwidth]{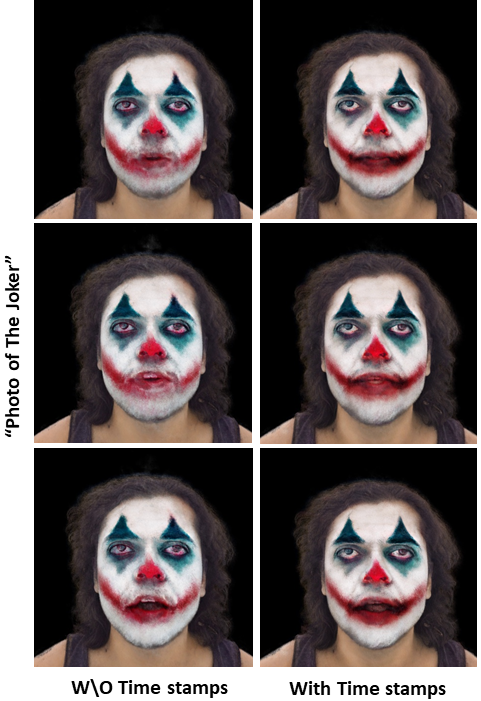}
    \caption{Fine-tuning LDM (Sec.~\ref{sec:FinetuneST}) without any temporal information leads to clear temporal inconsistencies (e.g. see mouth region).} 
    \label{fig:Kartik_time}
\end{figure}

We evaluate the various design choices of our method in an ablation study. 
First, we investigate the importance of our optimization strategy which accounts for multiple camera viewpoints and different time stamps during model fine-tuning (Sec.~\ref{sec:FinetuneST}),
To achieve this, we perform two experiments. First, we replace our optimization strategy with DreamBooth's~\cite{ruiz2022dreambooth}. While we still embed multiple camera viewpoints and different time stamps, 
however, DreamBooth's fine-tuning strategy only estimates one token for all images. Fig.~\ref{fig:importance of optimization} shows the output of this process. It is clear that this strategy leads to significant artifacts. 
However, since our approach estimates multiple tokens for each of the embedded frames, we can generates significantly better results (see top row). 
We also investigate the impact of not incorporating any time stamps in our fine-tuning. To achieve this, we used the same optimization discussed in Sec.~\ref{sec:FinetuneST}, 
however, we incorporated just the different camera viewpoints at time 0. 
Fig.~\ref{fig:Navami_time}-\ref{fig:Kartik_time} shows that this strategy leads to clear temporal inconsistencies in the output. Please see the supplemental video. 
This is a critical problem, especially during video edits. Incorporating different time stamps, however, clearly produces better results with temporally coherent output. 

\begin{figure*}
    \centering    \includegraphics[width=\textwidth]{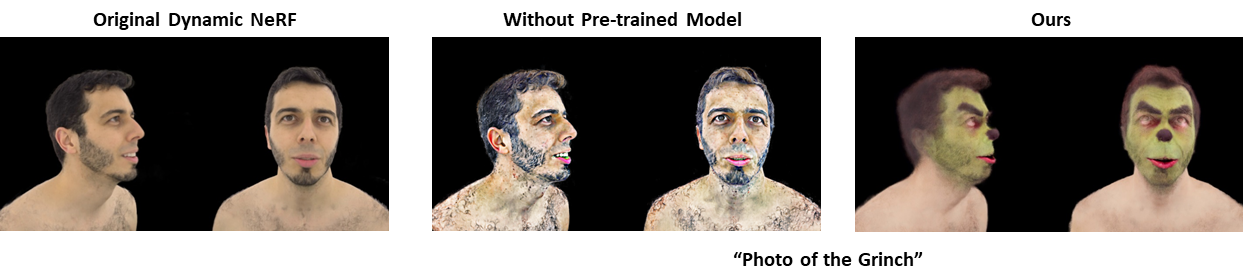}
    \caption{Removing the pre-trained model during Score Distillation Sampling lead to a significant drop in performance.} 
    \label{fig:SDSRatio}
\end{figure*}

\begin{figure*}
    \centering    \includegraphics[width=\textwidth]{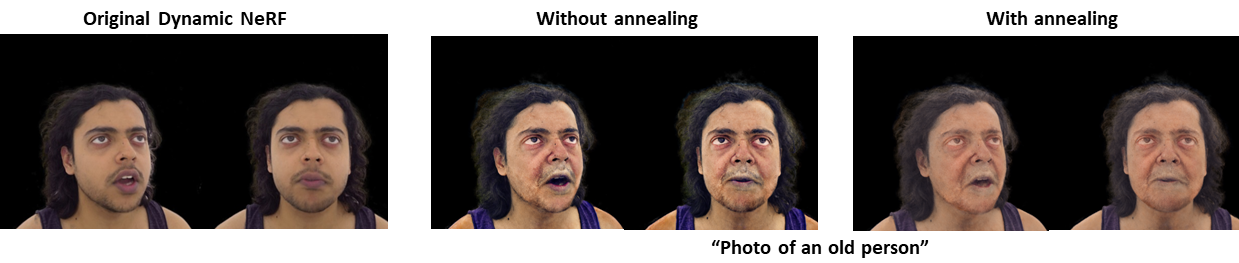}
    \caption{Using annealing during Score Distillation Sampling leads to better capturing of details.} 
    \label{fig:annealling}
\end{figure*}

The second part of our ablation study investigates the importance of our modified Score Distillation Sampling of Sec.~\ref{sec:editing}.
More specifically, we investigate the importance of using  Eq.~\ref{equ: modified classifier-free} during Score Distillation Sampling. 
To this end, we performed two main experiments. 
In the first experiment, we set v and K of Eq.~\ref{equ: modified classifier-free} to 0. This examines the importance of the pre-trained model during Score Distillation Sampling. 
Here, Fig.~\ref{fig:SDSRatio} shows that by removing the pre-trained model, results undergo a significant drop in performance.    
In the second experiment, we attempted to use the original SDS loss of DreamFusion~\cite{poole2022dreamfusion}. This approach, however, did not converge to any useful output. 
These experiments shows the importance of our modified Score Distillation Sampling.
Last but not lest, Fig.~\ref{fig:annealling} shows that using the SDS annealing of Lin~\etal~\cite{lin2022magic3d} leads to better details in the final results. 

\section{Discussion and Future Work}
We presented the first method for text-driven edits of dynamic head avatars. Our method utilizes a state-of-the-art NeRF based representation for dynamic heads and produces edits that are temporally coherent. Our results can be viewed from an arbitrary camera viewpoint in a 3D-consistent manner. At the heart of our method is a novel optimization strategy that incorporates multiple camera viewpoints, and multiple frames taken at different time stamps, in a pre-trained latent diffusion model. 
In addition, we proposed a new view-and-time-aware Score Distillation Sampling approach that combines knowledge from the pre-trained model, as well as our fine-tuned model. Our method enables a wide variety of text-driven edits and can produce
both photorealistic and non-photorealistic edits. We compared against related methods and results show that our approach produces better edits that are more temporally stable and more inline with the text prompts. 
This is confirmed visually, as well as numerically via a user study. 

Our work pushes the boundaries of text-driven visual edits. Nevertheless, several interesting avenues are still open for future work. 
While our method can produce a wide variety of edits, it requires multi-view data captured in a uniform illumination. 
Thus a very interesting research direction will be to handle just a monocular video as input, shot with in-the-wild conditions. 
This could be followed up, for instance, with a method that takes just a single image as input and rigs the output according to a target motion. 
While some of our results show geometrical edits, future work could look into producing edits that change the head geometry more drastically.
Currently our method is computationally expensive, requiring around 60 minutes train using a single A100 GPU. 
Future work could look into reducing this computational cost.
We hope that our work in text-driven visual editing encourages further research in this interesting problem.

\bibliographystyle{ACM-Reference-Format}
\bibliography{bibliography}

\end{document}